\documentclass[preprint,12pt%,double-blind
]{elsarticle}

\usepackage{amssymb}
\usepackage{amsmath}
\usepackage{url}
\usepackage{tikz}
\usepackage{cleveref}
\usepackage{multirow}
\usepackage{booktabs}
\usepackage{subcaption}
\usepackage{pgfplots}

\usetikzlibrary{fit, positioning}
\usepgfplotslibrary{groupplots}
\pgfplotsset{compat=newest}
\pgfplotsset{grid style={dashed,gray}}
\tikzset{fit margins/.style={/tikz/afit/.cd,#1,
    /tikz/.cd,
    inner xsep=\pgfkeysvalueof{/tikz/afit/left}+\pgfkeysvalueof{/tikz/afit/right},
    inner ysep=\pgfkeysvalueof{/tikz/afit/top}+\pgfkeysvalueof{/tikz/afit/bottom},
    xshift=-\pgfkeysvalueof{/tikz/afit/left}+\pgfkeysvalueof{/tikz/afit/right},
    yshift=-\pgfkeysvalueof{/tikz/afit/bottom}+\pgfkeysvalueof{/tikz/afit/top}},
    afit/.cd,left/.initial=2pt,right/.initial=2pt,bottom/.initial=2pt,top/.initial=2pt}
    
\DeclareRobustCommand{\bbone}{\text{\usefont{U}{bbold}{m}{n}1}}

\journal{International Journal of Forecasting}

\begin{document}

\begin{frontmatter}


\author[1]{Benedikt Heidrich\corref{cor1}} %\fnref{label2}
\ead{benedikt.heidrich@kit.edu}
%\fntext[label2]{}
\cortext[cor1]{Corresponding author.}
% \affiliation{organization={},

\author[1]{Kaleb Phipps}
\author[1]{Oliver Neumann}
\author[1]{Marian Turowski}
\author[1]{Ralf Mikut}
\author[1]{Veit Hagenmeyer}

\title{ProbPNN: Enhancing Deep Probabilistic Forecasting with Statistical Information}


\affiliation[1]{
    organization={Institute for Automation and Applied Informatics, Karlsruhe Institute of Technology},
    %addressline={Hermann-von-Helmholtz-Platz 1},
    city={Eggenstein-Leopoldshafen},
    postcode={76344},
    %state={Baden-Wuerttemberg},
    country={Germany}
}

\begin{abstract}
Probabilistic forecasts are essential for various downstream applications such as business development, traffic planning, and electrical grid balancing. 
Many of these probabilistic forecasts are performed on time series data that contain calendar-driven periodicities. 
However, existing probabilistic forecasting methods do not explicitly take these periodicities into account. Therefore, in the present paper, we introduce a deep learning-based method that considers these calendar-driven periodicities explicitly.
The present paper, thus, has a twofold contribution: First, we apply statistical methods that use calendar-driven prior knowledge to create rolling statistics and combine them with neural networks to provide better probabilistic forecasts. Second, we benchmark ProbPNN with state-of-the-art benchmarks by comparing the achieved normalised continuous ranked probability score (nCRPS) and normalised Pinball Loss (nPL) on two data sets containing in total more than 1000 time series. The results of the benchmarks show that using statistical forecasting components improves the probabilistic forecast performance and that ProbPNN outperforms other deep learning forecasting methods whilst requiring less computation costs.
\end{abstract}


\begin{highlights}
\item We combine statistical methods and deep learning-based forecasting methods to enhance probabilistic forecasts.
\item We evaluate ProbPNN empirically on more than 1000 time series from an Electricity and a Traffic data set. On these datasets, the proposed ProbPNN outperforms existing state-of-the-art methods.
\end{highlights}


\end{frontmatter}


\section{Introduction}\label{sec:introduction}
Accurate probabilistic forecasts are essential for various downstream applications and domains. E.g., in traffic planning, accurate probabilistic forecasts can support and improve traffic control or mitigate congestion \cite{Liu2021}. Likewise, in the energy domain, accurate probabilistic forecasts are essential for the smart grid and for stabilising the energy system \cite{dannecker2015}. \\
This need for probabilistic forecasts results in various existing types of time series forecasting methods in the literature. One well-known type of forecasting methods is e.g. statistical methods, such as the moving average, exponential smoothing, or ARIMA methods \cite{hyndman2018forecasting}. Another type of forecasting methods is deep learning-based methods such as N-BEATS \cite{oreshkin2019} and DeepAR \cite{Salinas2019}. \\ 
Although these methods are designed for time series in general, there may be additional prior knowledge for specific classes of time series that is ignored by these methods. One of these specific classes are time series that contain one or more calendar-driven periodicities as present in the exemplary use cases mentioned above. For example, traffic occupancy is higher on workday mornings and evenings than at night due to the rush hour. In the energy domain, more electricity is consumed during the day than at night, when most people sleep. These known periodicities are a key aspect of the time series structure, and their explicit inclusion in a probabilistic forecasting method could improve probabilistic forecast performance.\\
Therefore, the present paper introduces the Probabilistic Profile Neural Network (ProbPNN). ProbPNN applies statistical methods that use calendar-driven prior knowledge to create rolling statistics. These rolling statistics reflect the calendar-driven periodicities and are combined with a neural network to provide better probabilistic time series forecasts. In particular, ProbPNN uses a rolling average and a rolling standard deviation calculated on the time series grouped by the considered calendar-driven prior information. These statistics are combined with convolutional layers that extract trend and colourful noise information and aggregate them as well as further statistical information. \\
The main contribution of this paper is twofold. First, we introduce ProbPNN which combines statistical methods and deep learning to provide more accurate probabilistic forecasts. Second, we empirically evaluate ProbPNN on more than 100 electrical load time series and more than 900 traffic occupancy time series and compare these results with state-of-the-art deep learning forecasting methods. This comparison shows that ProbPNN provides better forecasts than state-of-the-art methods.\\
The remainder of the present paper is structured as follows. First, an overview of related work is provided in \Cref{sec:related}. Afterwards, in \Cref{sec:probpnn}, we introduce our proposed Probabilistic Profile Neural Network (ProbPNN). In the following \Cref{sec:eval}, we evaluate ProbPNN, and \Cref{sec:discussion} discusses the proposed ProbPNN and the results. Finally, in \Cref{sec:conclusion}, we wrap up the paper and provide an outlook on future research directions.

\section{Related Work}\label{sec:related}
In literature, numerous deep learning-based works exist for deterministic time series forecasting, including several standard architectures \cite{petropoulos2022forecasting} such as fully connected neural networks (FCNs) \cite{tealab2018time}, recurrent neural networks (RNNs) \cite{hewamalage2021}, and convolutional neural networks (CNNs) \cite{Koprinska2018}. There are also more complex methods. For example, the ES-RNN combines statistical methods like exponential smoothing with RNN \cite{Smyl2020}. Moreover, transformers are applied to time series as in the temporal fusion transformers (TFT) \cite{lim2021}, Informers \cite{Zhou2021}, and Autoformers \cite{Wu2021}. Furthermore, a novel architecture for time series forecasting is the neural decomposition-based network as in N-BEATS \cite{oreshkin2019}, NBEATSx \cite{Olivares22}, or NHiTS \cite{challu2022}. \\

However, these methods do not naturally provide probabilistic forecasts. 
%Whilst the previously mentioned neural networks represent the state-of-the-art in deterministic forecasting, they do not naturally create probabilistic forecasts.
To generate probabilistic forecasts based on neural network architectures three main strategies exist: \\
The first strategy aims to provide a probabilistic interpretation of a deterministic forecast. A popular method is to analyse the error of a forecast and assume that this error is normally distributed \cite{hyndman2018forecasting, petropoulos2022forecasting}. Based on this assumption, prediction intervals can be constructed that provide a probabilistic interpretation of the deterministic forecasts \cite{hyndman2018forecasting}.\\
The second strategy focuses on directly learning quantiles to represent the underlying probability distribution. Exemplary methods include using a specialised quantile loss function to train the neural network to learn the quantiles of a single time series \cite{koenker2017handbook, chung2021beyond} or a multivariate time series with the Multivariate Quantile Function Forecaster \cite{kan22}. Another way of learning quantiles is to determine them from the training data. For example, \cite{Ordiano2020} proposes a nearest neighbour quantile filter for determining the quantiles of each value in the training data. Afterwards, a neural network is trained to separately forecast these quantiles using a deterministic loss function. \\
The third strategy for obtaining probabilistic forecasts with a neural network is directly predicting the probability distribution parameters. Examples include the DeepAR network for univariate time series \cite{Flunkert2020} and the DeepVAR network for vectorised time series \cite{Salinas2019}. Both networks follow a similar idea and provide probabilistic forecasts by forecasting the parameters of a Gaussian distribution for each time step. \\
While these mentioned state-of-the-art methods concentrate on forecasting time series in general, the present paper focuses on time series with periodicities driven by calendar information. This focus allows our method to use specific prior knowledge to improve forecast accuracy. Additionally, in contrast to most state-of-the-art methods, our method integrates statistical methods into deep learning forecasting to improve probabilistic forecasts. 

\section{ProbPNN}\label{sec:probpnn}
The core idea of ProbPNN is that statistical features are beneficial for creating more accurate probabilistic forecasts. More specifically, ProbPNN extends previous work by using a rolling average (profile) and a rolling variance in combination with a special loss function that allows probabilistic forecasts to be made. \\ 
Therefore, this section first introduces how we calculate the rolling average and rolling variances and describes how we integrate prior knowledge in their calculation. Afterwards, we explain the architecture of ProbPNN, before describing its training.

\subsection{Rolling Average and Rolling Variance}
As mentioned before, ProbPNN uses statistical information to improve probabilistic forecasts. The statistical information used is the rolling average (i.e. a profile) and the rolling variance or rolling standard deviation. In the present paper, we incorporate existing calendar-driven prior knowledge into the calculation of this statistical information. Thus, the statistical information contains knowledge about calendar-driven periodicities. In particular, we statistically group the time series using calendar information before calculating each group's statistical information separately. For example, the profile or rolling variance value for Monday noon is calculated over the noon values of the previous working days. 
Formally, we calculate the moving average with
\begin{equation}
    p_{t} = \sum_{i=t-W}^{t-1} \frac{1}{n_{t, [t - W, t - 1]}}\begin{cases} l_i \text{,}  \quad & c(t, i) \\% \land \text{time\_of\_day}(t) = \text{time\_of\_day}(i) \\
            0\text{,}  \quad &\text{else},
            
    \end{cases} \label{eq:avg}
\end{equation}
where $t$ is the time for which the profile should be calculated, $c$ is a Boolean function that is true if the timestamps $t$ and $i$ belong to the same calendar-driven grouping\footnote{A calendar-driven grouping defines two timestamps with similar calendar-driven properties. For example, such a grouping may be defined by two timestamps with the same hour of the day. However, more complex groupings, such as two timestamps belonging to a weekend and having the same hour of the day, also form a calendar-driven grouping.}, $l_t$ is the value of the time series at point $t$, $W$ is the window length\footnote{We use $W=28$ days in the following.}, and $n_{t, [t - W, t - 1]}$ is the number of elements in $i \in [t - W, t - 1]$ with $c(t, i)$ equals true, i.e. the number of values in $[t - W, t - 1]$ that belong to the same calendar-driven grouping as $t$.
%The window length has to be set by the user (for the content of this paper, we set $W = 28$ days).

The calculation of the rolling variance is similar to the calculation of the moving average with
\begin{equation}
    v_{t} = \sum_{i=t-W}^{t-1} \frac{1}{n_{t, [t - W, t - 1]}}\begin{cases} (l_i - p_i)^2 \text{,}  \quad & c(t, i) \\
            0\text{,}  \quad &\text{else},
    \end{cases} \label{eq:var}
\end{equation}
with $t$, $c$, $l_t$, $W$, and $n_{t, [t - W, t - 1]}$ as defined above. To calculate the rolling standard deviation instead of the rolling variance, we calculate the square root of the rolling variance with
\begin{equation}
    \sigma_t = \sqrt{v_t}.
\end{equation}

\subsection{Network Architecture}
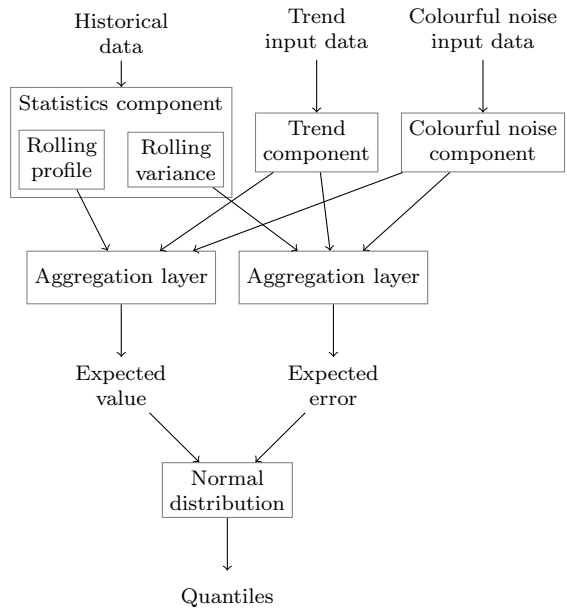
\begin{figure}
    \scriptsize
    \centering
        \begin{tikzpicture}[
	node distance=.7cm and .3cm, 
	minimum height=0.7cm,
	minimum width=0.8cm,
	title/.style={draw=none, color=gray, inner sep=0pt},
	typetag/.style={rectangle, draw=black!50, font=\footnotesize\ttfamily}
	]
	
	\node (mean) [draw=black!50, align=center] {Rolling\\profile};
	\node (var) [draw=black!50, align=center, right=of mean] {Rolling\\variance};
    \node (stats)[draw=black!50, align=center, fit=(mean)(var), fit margins={left=1.5pt,right=1.5pt,bottom=1.5pt,top=8pt}, label={[xshift=0.0cm, yshift=-0.6cm]Statistics component}] {};
    
	\node (trend) [draw=black!50, align=center,right= of stats] {Trend\\component};
	\node (noise) [draw=black!50, align=center,right= of trend] {Colourful noise\\component};
	\node (aggMu) [draw=black!50,align=center, below= of stats] {Aggregation layer};
	\node (aggSig) [draw=black!50,align=center, right= of aggMu] {Aggregation layer};
	\node (mu) [align=center,below= of aggMu] {Expected\\value};
	\node (sigma) [align=center, align=center, below= of aggSig] {Expected\\error};
	\node (histData) [align=center, align=center, above= of stats, yshift=-10] {Historical\\data};
	\node (trendData) [align=center, align=center, above= of trend] {Trend\\input data};
	\node (RemainderData) [align=center, align=center, above= of noise] {Colourful noise\\input data};
	\node (normal) [draw=black!50, align=center] at ($(mu)!.5!(sigma)+(0,-1.4cm)$) {Normal\\distribution};
	\node (quantiles) [align=center, below= of normal] {Quantiles};

        \draw[->] (histData) --  (stats);
        \draw[->] (trendData) --  (trend);
        \draw[->] (RemainderData) --  (noise);

        \draw[->] (mean) --  (aggMu);
        \draw[->] (trend) --  (aggMu);
        \draw[->] (noise) --  (aggMu);

        \draw[->] (var) --  (aggSig);
        \draw[->] (trend) --  (aggSig);
        \draw[->] (noise) --  (aggSig);
        
        \draw[->] (aggMu) --  (mu);
        \draw[->] (aggSig) --  (sigma);

        \draw[->] (mu) --  (normal);
        \draw[->] (sigma) --  (normal);

        \draw[->] (normal) --  (quantiles);

    \end{tikzpicture}
    
    \caption{The three components of ProbPNN are the statistics component, the trend component, and the colourful noise component. The statistics are extracted from the historical data, the trend data get the trend input data as input, and the colourful noise component gets the exogenous variables, and the difference between the profile and the historical data as input.
    Each component provides a mean and expected error for each value to be predicted. The aggregation layers then either combine the means or the expected errors to the forecast (expected value) or to the error of the forecast (expected error). The expected values and the expected error form the parameters of a normal distribution, from which the quantiles are derived.}
    \label{fig:pnn}
\end{figure}

The network architecture of ProbPNN is inspired by \cite{Heidrich2020} and comprises three components (see \Cref{fig:pnn}), namely the colourful noise component, the trend component, and the statistics component. In the following, we briefly introduce each of the components.

The colourful noise component consists of a historical data encoder and an exogenous feature encoder. The historical data encoder uses the last $k$ values of the time series as input. The exogenous feature encoder obtains the available exogenous features as input. As output, the colourful noise component provides the expected value of the colourful noise and the expected error of the colourful noise component for the forecast horizon. \Cref{tab:colorfulnoise} shows the architecture of the colourful noise network, including the historical data and the exogenous variables encoder, and \Cref{tab:subnet} the architecture of the convolutional subnetwork.  \\

% \usepackage{multirow}
\begin{table}
\centering
\scriptsize
\caption{The colourful noise component encodes the historical data and the exogenous variables. Afterwards, the encoded historical data and the encoded exogenous variables are concatenated on a new axis. A further convolutional network encodes the concatenated data. The flattened output of this convolutional network is finally used by two concurrent dense layers to forecast the mean and the variance or standard deviation.}\label{tab:colorfulnoise}
% \usepackage{multirow}
\begin{tabular}{cc}
\toprule
\multicolumn{1}{c|}{Historical Data} &  Exogenous Variables \\ \midrule
\multicolumn{1}{c|}{ConvNet} & ConvNet \\
\multicolumn{1}{c|}{Dense horizon} & \\ \midrule
\multicolumn{2}{c}{Concatenate on a new axis}                                                                                                                                                                                                                                \\
\multicolumn{2}{c}{ConvNet}                                                                                                                                                                                                                                   \\  
\multicolumn{2}{c}{Flatten}                                                                                                                                                                                                                                   \\  \midrule
\multicolumn{1}{c|}{Dense 24 Activation Linear}                                                                                                     & \multicolumn{1}{c}{Dense 24 Activation Linear}      \\  \midrule
\multicolumn{1}{c|}{Expected Value}                                                                                                     & \multicolumn{1}{c}{Expected Error}      \\  
\bottomrule
\end{tabular}
\end{table}

\begin{table}
    \caption{The convolutional subnetwork used in the colourful noise component and the trend component consists of six one dimensional convolution layers with varying number of filters and either the exponential linear unit (Elu) or the linear activation function . The kernel size of each convolution is three.}
    \label{tab:subnet}
    \centering
    \scriptsize
    \begin{tabular}{cl} \toprule
         Layer & Description \\ \midrule
         1 & 1D Convolution with 4 filters and Elu \\
         2 & 1D Convolution with 8 filters and Elu \\
         3 & 1D Convolution with 16 filters and Elu\\
         4 & 1D Convolution with 32 filters and Elu \\
         5 & 1D Convolution with 1 filters and Linear
          \\ \bottomrule
    \end{tabular}
\end{table}

The second component of ProbPNN is the trend component. For each value to be predicted, the trend component takes the last $m$ values that match the same periodicity. For example, for hourly electrical load data with weekly periodicities, the trend component considers the values $l_{\hat{t}-s\cdot m}, l_{\hat{t}-s\cdot (m - 1)},\dots, l_{\hat{t}-s}$, where $s$ is the length of one periodicity\footnote{We use $s=168$ in the following to consider a weekly periodicity and an hourly resolution.}, and $\hat{t} \in [t +1, t+\operatorname{horizon}]$ are the timesteps of the values to be predicted. The output of the trend component is the expected value and the expected error of this component. 
The trend component consists of a convolutional network and two dense layers for forecasting the expected value and the expected error. The \Cref{tab:trend} shows the architecture of the trend component.

\begin{table}
    \centering
    \caption{The trend component consists of one convolutional subnetwork that encodes the trend input data. The flattened output of this convolutional network is afterwards used by two concurrent dense layers to forecast the mean and the variance or standard deviation.}
    \label{tab:trend}
    \centering
    \scriptsize
    \begin{tabular}{c|c} \toprule
         \multicolumn{2}{c}{Trend input data}  \\ \midrule
         \multicolumn{2}{c}{Conv Net} \\
         \multicolumn{2}{c}{Flatten}\\ \midrule 
         Dense & Dense \\ 
         Expected value & Expected Error\\ \bottomrule
    \end{tabular}
\end{table}

The third component of ProbPNN is the statistics component. The statistics component compromises the rolling average and the rolling variance. 
Both statistics are directly used in ProbPNN without any encoding layers.
\\

Finally, the output of the three components is weighted via an aggregation layer. It multiplies the output of each component by a weight and sums these weighted outputs together. The weights are determined in the training process of ProbPNN. The final output is the forecasted time series $\hat{y}$ and the expected error of the forecast $\hat{e}$.

\subsection{Training ProbPNN}

In the training process, ProbPNN learns two target variables, namely the expected value and the expected error of the forecast. Since both have to be considered in the loss, we first describe a loss function that takes into account the expected value. Afterwards, we explain how a loss function can consider the estimated error. Finally, we describe how we combine these two loss functions.

To learn the expected value of the time series, we use the mean absolute error (MAE) as loss function. It is defined as  
\begin{equation}
    L_1 = \operatorname{MAE}(\hat{y}, y) = \frac{\sum_{i=1}^n \mid \hat{y}_i - y_i \mid}{n}
\end{equation}
where $\hat{y}$ is the forecast value, $y$ is the actual value, and $n$ is the length of the sample\footnote{Alternatively, other losses such as the Mean Squared Error (MSE) can be used, too.}.

Regarding the expected error of the forecast, we distinguish two cases. In the first case, the expected error is measured by an absolute error, whilst in the second case, this expected error is measured by a squared error. The first case should be applied when the estimated error corresponds to the variance of the forecast. Thus, we call the resulting model ProbPNN-$\sigma^2$. The second case must be applied when the learned error corresponds to the standard deviation. The corresponding model is called ProbPNN-$\sigma$.
To reflect both cases, we use the loss function
\begin{equation}
    L_2 = \begin{cases}
        \frac{1}{n} \sum_{i=1}^n \mid (\hat{y}_i - y_i)^2 - \hat{e} \mid & \text{ProbPNN-}\sigma^2, \\
        \frac{1}{n} \sum_{i=1}^n \mid (\mid \hat{y}_i - y_i \mid) - \hat{e} \mid & \text{ProbPNN-}\sigma 
    \end{cases}
\end{equation}
where $n$ is the number of samples in the batch, $\hat{y}$ is the forecast value of the time series, $y$ is the actual value of the time series, and $\hat{e}$ is the estimated error of the forecast value $\hat{y}$. 
Note that the selection of the $L_2$ loss influences which statistical component is required: If the ProbPNN-$\sigma$ should be learned, the rolling standard deviation has to be used in the statistics component. If the ProbPNN-$\sigma^2$ should be learned, the rolling variance must be used in the statistics component. %In the following, we refer to ProbPNN that learns the standard deviation and uses the MAE-based $L_2$ as ProbPNN-$\sigma$ and the PNN that learns the variance and uses the MSE-based $L_2$ as ProbPNN-$\sigma^2$.
%If the standard deviation should be learned, an MAE-based loss has to be selected, if the variance should be learned an MSE-based loss has to be selected. \footnote{Note, if you aim to learn the standard deviation, you should input the rolling standard deviation in the network instead of the rolling variance}. \\
%E.g., with the MAE, ProbPNN will learn the standard deviation and with the MSE the variance.

Finally, we combine $L_1$ and $L_2$ to form the final loss $L$. We use adaptive weights to ensure that both losses are considered equally important in the training process. These weights are calculated as

\begin{equation}
    \begin{split}
        w_1 = \frac{L_2}{L_1 + L_2},
        w_2 = \frac{L_1}{L_1 + L_2}. \\
    \end{split}
\end{equation}

In the final loss, the weight $w_1$ is multiplied by the loss $L_1$ and $w_2$ is multiplied by the loss $L_2$, resulting in the final loss function
\begin{equation}
    L = w_1 \cdot L_1 + w_2 \cdot L_2.
\end{equation}

\section{Evaluation}\label{sec:eval}
To comprehensively evaluate ProbPNN, we evaluate both variants of ProbPNN: ProbPNN-$\sigma$ and ProbPNN-$\sigma^2$. Both are implemented in Keras \cite{Chollet2015} and TensorFlow \cite{Abadi2015}. To evaluate and compare ProbPNN with the benchmarks, we build a pipeline using pyWATTS
\footnote{\url{https://github.com/KIT-IAI/pyWATTS}
} \cite{Heidrich2021}.
\footnote{The implementation will be made available after the acceptance of the paper.}
We evaluate ProbPNN-$\sigma$ and ProbPNN-$\sigma^2$ on two publicly available data sets, which we describe in the following subsection. Afterwards, we introduce the used benchmarks and metrics. Finally, we present the evaluation results.

\subsection{Data}
For the evaluation, we use two publicly available data sets from the UCI repository \cite{Dua2019} that are also used in e.g.~\cite{Flunkert2020, Yu2016}. \Cref{tab:datasets} provides an overview of both data sets.

The first data set is the Electricity data set\footnote{\url{https://archive.ics.uci.edu/ml/datasets/ElectricityLoadDiagrams20112014}}. It comprises time series representing the electrical load of several clients in Portugal. Since multiple time series contain long periods of almost constant values, we use a subset for our evaluation. We select all clients that have less than eight hours of consecutive zeros or not almost constant values over several days, since we assume that such values are related to measurement errors or non-available data.\\ 
%with a quarter-hourly resolution from the beginning of 2011 until the end of 2014 \cite{Rodrigues2018}. For the evaluation, we aggregate the dataset to an hourly resolution. Furthermore, since multiple time series contain long periods of zeros, we select as a subset the electrical load of 130 consumers. As criterium for the subset, we select all buildings that have less than 8 hours of consecutive zeros, since we assume that more are related to measurement errors or non-available data. When using this dataset, we train on the first three years and use the last year for testing. \\
%
The second data set is the Traffic data set\footnote{\url{https://archive.ics.uci.edu/ml/datasets/PEMS-SF}}. This data set contains time series representing the occupancy rate of car lanes in the San Francisco Bay Area. We use all time series from this data set for our evaluation.
%This data set contains measurements of the occupancy rate of 963 car lanes in the San Francisco Bay Area. This dataset spans from January 2008 to March 2009 with a ten-minute resolution. Analog to the Electricity dataset, we aggregate this dataset also to an hourly resolution. In contrast, to the Electricity dataset, we use every time series from this dataset for the evaluation. When using this dataset, we train the models on the first year of the data and use the three months available in 2009 for testing.

To ensure a fair comparison of ProbPNN and the benchmarks, we provide all methods with the same input features. As historical data, each method receives either the past 36 hours of data or the past 36 hours of difference between the data and the rolling average. Moreover, each method receives a rolling profile and a rolling variance as inputs. Furthermore, all methods obtain exogenous features that differ between the two data sets. 
%For the electricity data set, we use the sine- and cosine-encoded time of the day, the sine- and cosine-encoded month of the year, and a Boolean that indicates whether the current day is a weekend day or not. For the traffic data set, we use the sine- and cosine-encoded time of the day and a Boolean indicating whether the current day is a weekend day or not. 

\begin{table*}
    \centering
    \scriptsize
    \caption{Overview of the data sets used for the evaluation.}
    \label{tab:datasets}
    \begin{tabular}{p{4cm}p{4cm}p{4cm}} \toprule
         &  Electricity & Traffic \\ \midrule
         Time period & January 2011 -- December 2014 & January 2008 -- March 2009 \\
         Number of time series & 370 & 963\\
         Number of used time series & 130 & 963 \\
         Original resolution & 15 minutes & 10 minutes \\
         Used resolution & Hourly & Hourly \\
         Train set & January 2011 -- December 2013 &  January 2008 -- December 2008\\
         Test set & January 2014 -- December 2014 & January 2009 -- March 2009 \\ 
         Exogenous variables & Sine/cosine-encoded time of the day and month of the year and a Flag for weekends & Sine/cosine-encoded time of the day and a Flag for weekends \\ \bottomrule
    \end{tabular}
\end{table*}

\subsection{Benchmarks}
To assess the quality of ProbPNN's probabilistic forecast, we compare it with benchmarks. For this comparison, we select six probabilistic forecasts.  \\
The first benchmark is a purely statistical forecast called profile-standard deviation forecast (PSF), which serves as a baseline. The PSF calculates the rolling average (\Cref{eq:avg}) and the rolling standard deviation (\Cref{eq:var}) on a window with a size of 28 days. To calculate these two statistics, we statistically group the time series by the hour of the day.
The resulting rolling average and variance are then interpreted as parameters of a Gaussian distribution leading to a probabilistic forecast.  \\
%The second forecast is the Linear Regression plus Prediction Interval (LRPI) Forecast. It is based on \cite{Heidrich2022}. It takes a Linear Regression to predict the difference between a rolling profile and the actual consumption. On basis of this forecast a Prediction Interval is used to get probabilistic forecasts. \todo{Rausnehmen?} \\
The other five benchmarks are neural network-based. They cover the previously described three strategies of probabilistic forecasts with neural networks.
The first neural network-based benchmark is the PNN \cite{Heidrich2020} combined with prediction intervals (PNN-PI) to create a probabilistic interpretation of the forecast. For the PNN-PI, we use its implementation in pyWATTS \cite{Heidrich2021}.
%The second neural network based forecast is a Quantile Regression Neural Network (QRNN), which predicts the 99 quantiles from 1 to 99. \todo{rausnehmen? Ist ja im Prinzip wie NHiTS und TFT}.
The second and third neural network-based benchmarks are the state-of-the-art forecasting networks temporal fusion transformers (TFT) \cite{lim2021} and NHiTS \cite{challu2022} equipped with quantile loss functions. For both benchmarks, we use the implementations provided by PyTorch Forecasting\footnote{https://pytorch-forecasting.readthedocs.io/en/stable/}. 
The fourth neural network-based benchmark uses the nearest neighbour quantile filter (NNQF) proposed in \cite{Ordiano2020} for providing probabilistic forecasts. It is a simple MLPRegressor from sklearn \cite{Pedregosa2011} that directly predicts the quantiles that are previously extracted by a nearest neighbour quantile filter.
The last neural network-based benchmark is DeepAR \cite{Flunkert2020}, which is also a state-of-the-art forecasting network that directly predicts a distribution as a probabilistic forecast. For DeepAR, we also use the implementation provided by PyTorch Forecasting.

\subsection{Evaluation Metrics}
To evaluate ProbPNN and compare it with the benchmarks, we use five evaluation metrics: The normalised continuous ranked probability score (nCRPS) and normalised pinball loss (nPL) assess the quality of the probabilistic forecast. The normalised mean absolute error (nMAE) and the distance to the ideal coverage rate (DICR) provide insights into the accuracy of the median forecast and the probability distribution. The training time assesses the computational effort. In the following, we briefly introduce each of the five metrics.

\paragraph{Normalised Continuous Ranked Probability Score}
To assess the quality of our probabilistic forecast, we use the normalised continuous ranked probability score (nCRPS) \cite{matheson1976scoring}. It is defined as 
\begin{equation}
    \operatorname{nCRPS} (F, y) = \frac{1}{y_{\text{max}}} \int_\mathcal{R} (F(x) - 1(x \geq y))^2 dx,
\end{equation}
where $F$ is the forecast cumulative distribution function, $y$ the actual value, and $y_{\text{max}}$ is the maximum of the considered time series. To calculate the nCRPS, we use the ensemble-based implementation provided by the \texttt{properscoring} python package\footnote{https://github.com/properscoring/properscoring}.

\paragraph{Normalised Pinball Loss}
The normalised pinball loss (nPL) assesses the quality of a specific quantile forecast. To use the nPL for a probabilistic forecast, we calculate the average nPL over all quantiles of the probabilistic forecast. It is thus defined as
\begin{equation}
    \operatorname{nPL} = \frac{1}{y_{\text{max}} \cdot n \cdot \mid Q \mid}\sum_{\alpha \in Q} \sum\limits_{i=1}^n \begin{cases} (y_i - \hat{y}_{\alpha, i}) \cdot \alpha \quad  y_i \geq \hat{y}_{\alpha_i} \\
    (\hat{y}_{\alpha,i }-y_i) \cdot (1-\alpha) \quad \hat{y}_{\alpha, i} > y,
        
    \end{cases}
\end{equation}
where $y$ is the actual value, $\hat{y}_\alpha$ is the quantile forecast for the quantile $\alpha$, $Q$ is the set of all considered quantiles\footnote{For the evaluation, we use $Q=\{0.99, 0.98, 0.97, 0.96, 0.95, 0.9, 0.85, ..., 0.05, 0.04, 0.03,$$ 0.02, 0.01\}$.}, $\mid Q \mid $ is the cardinality of $Q$, and $y_{\text{max}}$ is the maximum of the considered time series. 

\paragraph{Distance to the Ideal Coverage Rate}
To also consider the accuracy of the quantiles, we use the coverage rate (CR). It measures the share of actual values that lie in the interval between two quantiles, i.e.
\begin{equation} \label{eq:cr}
    \operatorname{CR}_{q_1, q_2} = \frac{1}{n}\sum_{i=1}^n \bbone(q_1 < y_i < q_2)
\end{equation}
where $q_1$ and $q_2$ are the lower and upper considered quantiles, $y$ is the observed value, $\bbone(q_1 < y < q_2)$ is the indicator function that counts how many values are between $q_1$ and $q_2$, and $n$ is the number of observations. Since an ideal forecast would result in a CR of $q_2 - q_1$, we adapt this metric as follows. We use the sum of the distances of a set of quantile tuples to the ideal forecast to form the distance to the ideal coverage rate (DICR), i.e.,
\begin{equation}
    \operatorname{DICR} = \sum_{q_i, q_j \in Q} \mid (CR_{q_i, q_j} - (q_j - q_i)) \mid ,
\end{equation}
where $Q$\footnote{For the evaluation, we use $Q = \{(0.05, 0.95), (0.1, 0.9), ... (0.45, 0.55)\}$.} is the set of all considered quantile tuples and $q_j > q_i$.

\paragraph{Normalised Mean Absolute Error}
In addition to the probabilistic forecasts, we also evaluate the accuracy of the point forecast derived from the probabilistic forecast. To obtain a point forecast, we assess the median using the normalised mean absolute error (nMAE). It is defined as
\begin{equation}
    \operatorname{nMAE} = \frac{1}{y_{\text{max}} \cdot n} \sum_{i=1}^n \mid  \hat{y}_i - y_i \mid,
\end{equation}
where $n$ is the length of the test data, $\hat{y}$ the forecast value, $y$ the actual value, and $y_{\text{max}}$ is the maximum of the considered time series.

\paragraph{Training Time}
In addition to evaluating the forecast accuracy, we also examine the training time of ProbPNN and the benchmarks. For this, we measure the needed training time of the introduced methods in seconds (see \Cref{tab:hardware} for the used hardware). 

\begin{table}
    \centering
    \scriptsize
    \caption{The hardware setup used for the evaluation. Note that, due to the training times and the more than 1000 considered time series, we use two computers for the evaluation on the two data sets.}
    \label{tab:hardware}
    \begin{tabular}{lll} \toprule
               &  Electricity & Traffic \\ \midrule
    CPU cores  & 96 & 32 \\
    RAM & 126GB & 64GB \\
    GPU & no GPU & Nvidia Titan RTX 24GB \\ \bottomrule
    \end{tabular}
\end{table}

\subsection{Results}
Given the data, benchmarks, and evaluation metrics introduced above, we present the results in the following.
First, we benchmark ProbPNN-$\sigma$ and ProbPNN-$\sigma^2$ by comparing their probabilistic forecast quality to the benchmarks. Second, we investigate the resulting distance to the ideal coverage rates and the derived deterministic forecast of both ProbPNN-$\sigma$ and ProbPNN-$\sigma^2$ and the benchmarks to gain insights. Third, we compare the training time of both ProbPNN-$\sigma$ and ProbPNN-$\sigma^2$ and the benchmarks.

\subsubsection{Benchmarking}

To compare ProbPNN and the benchmarks on the Electricity and Traffic data sets regarding the quality of the probabilistic forecast, we calculate the average nCRPS and the average nPL of all considered time series from the Electricity and Traffic data sets, respectively.
In addition to the average nCRPS and nPL, we also calculate the average rank of each method of all considered time series for each data set.

\Cref{tab:benchmarking} reports the scores and ranks for the nCRPS and nPL for all evaluated methods and both data sets. In the results, we make three observations.
First, we observe that the proposed ProbPNN outperforms the benchmarks concerning both scores. Regarding the electricity data set, ProbPNN's nCRPS score is 6.8\% and nPL score is 4.5\% better than those of the TFT as the second best method. On the Traffic data set, the improvements are even larger. ProbPNN reduces the nCRPS score by 21.1\% and the nPL by 21.8\% compared to the TFT as the second-best method. \\
Second, with regard to the average rank, we also observe that ProbPNN-$\sigma$ outperforms the benchmarks clearly. On the Electricity data set, ProbPNN-$\sigma$ achieves a rank of 1.44 for the nCRPS and 1.39 for the nPL. TFT as the best benchmark obtains an average rank of 2.76 for the nCRPS and 2.84 for the nPL. On the Traffic data set, ProbPNN-$\sigma$ has a rank of 1.38 for the nCRPS and 1.34 for the nPL, while the TFT's rank is 3.2 for the nCRPS and 2.99 for the nPL. \\
Third, when comparing ProbPNN-$\sigma$ and ProbPNN-$\sigma^2$, we observe that ProbPNN-$\sigma$'s scores and ranks are equal to or lower than those of ProbPNN-$\sigma^2$ on both data sets for both metrics. \\

% \usepackage{multirow}
\begin{table*}[t]
    \caption{Probabilistic forecasting quality of ProbPNN-$\sigma$ (PPNN-$\sigma$) and ProbPNN-$\sigma^2$ (PPNN-$\sigma^2$) as well as the simple profile and standard deviation-based forecast (PSF), the Profile Neural Network with Prediction Intervals (PNN-PI), the Temporal Fusion Transformers (TFT), the NHiTS, the Nearest Neighbour Quantile Filter with an MLP as a regressor (NNQF), and DeepAR for both data sets.}
    \label{tab:benchmarking}
    \begin{subtable}[t]{\textwidth}
    \centering
    \scriptsize
    \caption{Average nCRPS scores and ranks of all considered time series from the Electricity respectively Traffic data set.}
    \begin{tabular}{llllllllll} \toprule
                                 &       & PPNN-$\sigma$ & PPNN-$\sigma^2$ & PSF    & PNN-PI  & TFT    & NHiTS  & NNQF  & DeepAR    \\ \midrule
    \multirow{2}{*}{Electricity} & Score & 0.041  & 0.041  & 0.050  & 0.054  & 0.044  & 0.058  & 0.053  & 0.049  \\
                                 & Rank  & 1.44    & 1.90    & 5.28        & 6.33   & 2.76 & 7.37  & 6.44 & 4.48        \\ \midrule
    \multirow{2}{*}{Traffic}     & Score & 0.086   & 0.098   & 0.185       & 0.197  & 0.109 & 0.139 & 0.138 & 0.139           \\
                                 & Rank  &  1.38 & 2.52 & 8.03 & 7.98 & 3.2 & 5.83 & 5.47 & 5.68      \\ \bottomrule
    \end{tabular}
    \end{subtable}
    
    % \usepackage{multirow}
    \begin{subtable}[t]{\textwidth}
    \scriptsize
    \centering
    \caption{Average nPLs scores and ranks of all considered time series from the Electricity respectively Traffic data set.}
    \begin{tabular}{llllllllll} \toprule
                                 &       & PPNN-$\sigma$ & PPNN-$\sigma^2$ & PSF   & PNN-PI & TFT   & NHiTS & NNQF & DeepAR   \\\midrule
    \multirow{2}{*}{Electricity} & Score & 0.021  & 0.021  & 0.025 & 0.027 & 0.022 & 0.029 & 0.027 & 0.025 \\
                                 & Rank  & 1.39    & 1.86    & 5.27        & 6.34   & 2.84 & 7.41  & 6.42 & 4.48    \\ \midrule
    \multirow{2}{*}{Traffic}     & Score & 0.043   & 0.049   & 0.093       & 0.100  & 0.055 & 0.071 & 0.071 & 0.070   \\
                                 & Rank  & 1.34    & 2.42    & 6.97        & 7.04   & 2.99  & 5.29  & 4.93  & 5.02 \\ \bottomrule
    \end{tabular}
    \end{subtable}
\end{table*}

\subsubsection{Insights}
\begin{figure*}
    \centering
    \scriptsize
    \input{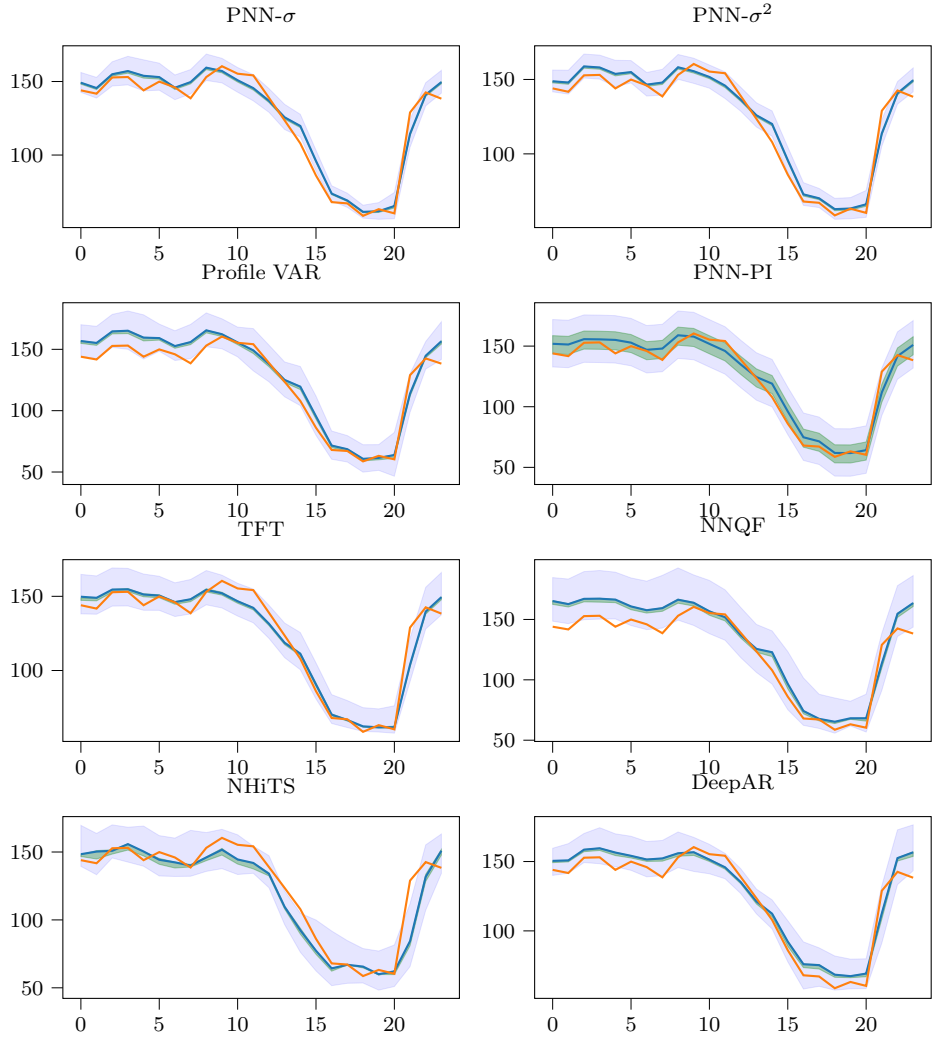}
    \caption{Ground truth (orange), forecast (dark blue), the $90\%$ (light blue) and $20\%$ (green) confidence intervals of ProbPNN-$\sigma$, ProbPNN-$\sigma^2$, the simple profile and standard deviation-based forecast (PSF), the Profile Neural Network with Prediction Intervals (PNN-PI), the Temporal Fusion Transformers (TFT), the NHiTS, the Nearest Neighbour Quantile Regression Filter with an MLP as a regressor (NNQF), and DeepAR for an exemplary day of the electricity data set.}
    \label{fig:forecast}
\end{figure*}

%To determine whether ProbPNN's performance is mainly driven by an improved probability distribution or an improved deterministic forecast, we investigate ProbPNN in more detail.
To gain further insights into the performance of the proposed ProbPNN, we also investigate ProbPNN regarding the resulting probability distribution and the derived deterministic forecast. 
For this, we perform two analyses: We analyse the difference between the achieved coverage rate and the ideal coverage rate. Additionally, we examine the accuracy of the median forecasts using the nMAE. Finally, we also provide a visualisation of an exemplary forecast.  \\
%Based on these analyses, we are able 
\Cref{tab:insights} shows the DICR and the nMAE for all evaluated methods and both data sets.
In the table, we make three observations. First, ProbPNN has the lowest DICR and the TFT the second lowest, whereas the PNN with Prediction Intervals has the highest CR. Second, in contrast to the previous observation, the PNN with Prediction Intervals slightly outperforms ProbPNN with respect to the accuracy of the deterministic forecast. Third, we observe that all methods generally obtain a high CR. 
%Finally, in \Cref{fig:forecast}, we also examine an exemplary forecast of the electricity dataset.
The exemplary forecast in \Cref{fig:forecast} confirms this observation since the shown quantile bands are small. 

% \begin{itemize}
%     \item CR: ProbPNN has the lowest distance to an ideal coverage rate, followed by the TFT. The worst CR is achieved by the PNN with Prediction Intervals. 
%     \item MAE: In contrast, regarding the MAE, PNN with PI achieves the best results and slightly outperforms ProbPNN with MAE. Indicating that the Predicition Intervals lead to worse Quantile Forecasts. 
%     \item However, in general, we observe that the distance to the optimal Coverate Rate of all methods is quiet high. \todo{Percentage deviation?}. This seems to be confirmed by the plot, where the Prediction Intervals are very small.
% \end{itemize}

\begin{table*}
\scriptsize
\centering
\caption{
The average DICR and the average nMAE of ProbPNN-$\sigma$ (PPNN-$\sigma$) and ProbPNN-$\sigma^2$ (PPNN-$\sigma^2$) as well as the simple profile and standard deviation-based forecast (PSF), the Profile Neural Network with Prediction Intervals (PNN-PI), the Temporal Fusion Transformers (TFT), the NHiTS, the Nearest Neighbour Quantile Regression Filter with an MLP as a regressor (NNQF), and DeepAR for both data sets.
}  \label{tab:insights}
\begin{tabular}{llllllllll} \toprule
                             & & PPNN-$\sigma$ & PPNN-$\sigma^2$ & PSF   & PNN-PI & TFT   & NHiTS & NNQF & DeepAR   \\ \midrule
\multirow{2}{*}{Electricity} & \begin{tabular}[c]{@{}l@{}}DICR\end{tabular} & 1.77   & 1.81   & 2.63  & 3.98  & 2.61  & 2.52  & 2.91  & 2.93  \\
                             & nMAE                                                     & 0.051  & 0.051  & 0.065 & 0.051 & 0.056 & 0.076 & 0.069 & 0.064 \\ \midrule
\multirow{2}{*}{Traffic}     & \begin{tabular}[c]{@{}l@{}}DICR\end{tabular} & 1.83    & 1.67    & 3.08        & 4.22   & 2.23 & 2.42  & 2.76 & 2.73             \\
& nMAE             & 0.11    & 0.12    & 0.24        & 0.11   & 0.14 & 0.18  & 0.18 & 0.19    \\ \bottomrule
\end{tabular}
\end{table*}

\subsubsection{Training Time}
To analyse the training time, we compare the training time required by ProbPNN and the benchmarks. \\

\Cref{tab:time} presents the training time in seconds needed by ProbPNN and the benchmarks for both data sets.
Based on the results, we describe two observations. First, we see that ProbPNN needs less training time than the state-of-the-art deep learning benchmarks and more than the NNQF. Note that the training time of the NNQF strongly depends on the used base learner. Second, we observe that ProbPNN-$\sigma$ has a higher training time than ProbPNN-$\sigma^2$.

\begin{table*}
    \centering
    \scriptsize
    \caption{Average training time in seconds of ProbPNN-$\sigma$ (PPNN-$\sigma$) and ProbPNN-$\sigma^2$ (PPNN-$\sigma^2$) as well as the Profile Neural Network with Prediction Intervals (PNN-PI), the Temporal Fusion Transformers (TFT), NHiTS, the Nearest Neighbour Quantile Regression Filter with an MLP as a regressor (NNQF), and DeepAR for both data sets. Note that the simple profile and standard deviation-based forecast (PSF) is omitted as it does not need to be trained.}
    \label{tab:time}
\begin{tabular}{llllllll} \toprule
            & PPNN-$\sigma$ & PPNN-$\sigma^2$ & PNN-PI & TFT  & NHiTS & NNQF & DeepAR  \\ \midrule
Electricity & 197    & 164    & 150   & 2995 & 549   & 79    & 2221 \\
Traffic     & 717     & 451     & 677    & 4762 & 914   & 50   & 2510    \\ \bottomrule
    \end{tabular}
\end{table*}

\section{Discussion}\label{sec:discussion}
In this section, we discuss the results, the limitations of the proposed ProbPNN, and its benefits. \\

The results of the benchmark show that ProbPNN outperforms the benchmarks on both selected data sets. We also note that ProbPNN-$\sigma$ provides better results than ProbPNN-$\sigma^2$. 
With regard to the distance to an ideal coverage rate, we observe for all evaluated methods that the distance to the ideal quantiles is rather high and thus the uncertainty is over- or underestimated. Adapted loss functions or additional post-processing strategies might improve the coverage rates. 
%investigating how the distance can be reduced for the evaluated methods, e.g. through better-suited loss functions or appropriate post-processing strategies.
Concerning the training time, our results show that the training time of ProbPNN is advantageously small compared to the deep learning benchmarks.
%This small training time is especially beneficial for training forecasting methods for individual time series at scale. 
%Nevertheless, the used implementations of the evaluated methods differ and might distort the results. Therefore, future work could comprehensively examine the computational effort of the evaluated methods by reimplementing them consistently. 

%Regarding the results, we discuss ... \\
%\begin{itemize}
%    \item ProbPNN outperforms the other benchmarks.
%    \item The confidences provided by ProbPNN are too small. Thus, future work should investigate how this can be improved. However, all methods that uses no postprocessing seems to suffer from that.
%    \item The computational effort seems to be beneficial for ProbPNN compared to the other deep learning approaches. However, the library used for implementing ProbPNN differs from the libraries used by the benchmarks. Thus, these results may change if ProbPNN and the benchmarks are implemented in the same library.
%\end{itemize}

Regarding the limitations of ProbPNN, we see three aspects worth discussing. 
First, ProbPNN exploits regular periodicities in time series to improve probabilistic forecasts. For this reason, ProbPNN's application is designed for time series containing calendar-driven periodicities. Second, ProbPNN assumes that the time series follows a normal distribution. On the tested data sets, this assumption does not seem to influence the results negatively since ProbPNN outperforms the benchmarks. However, time series are generally not always normally distributed, which could be addressed by an extension of ProbPNN. %Therefore, future work could investigate how to extend ProbPNN accordingly. 
Third, ProbPNN assumes the independence of the trend, colourful noise, and rolling statistics so that these three components can be summed together. This limitations can be addressed by applying copulas in ProbPNN.

Overall, ProbPNN enhances probabilistic forecasts using statistical information while requiring a small training time compared to other deep learning forecasting methods.

\section{Conclusion}\label{sec:conclusion}
In the present paper, we propose a novel probabilistic deep learning method for time series forecasting called ProbPNN. ProbPNN combines statistical components with deep learning and uses the prior knowledge contained in time series with calendar-driven periodicities. We evaluate ProbPNN on more than 1000 time series from two different data sets with calendar-driven periodicities.
The results show that the focus on time series with calendar-driven periodicities and the combination of statistical components and deep learning methods enable ProbPNN to outperform state-of-the-art forecasting methods on the tested data sets in probabilistic forecasting. \\

Future work may relax the assumption that the trend, colourful noise, and variance are independent by extending ProbPNN's aggregation layer with copulas. Moreover, it could be interesting to investigate how adapted loss functions or post-processing strategies influence the coverage rates.
Furthermore, ProbPNN could be extended to multivariate time series forecasting and to mitigate the assumption of not normally distributed errors.

\section*{Acknowledgements}
This project is funded by the Helmholtz Association’s Initiative and Networking Fund through Helmholtz AI and the Helmholtz Association under the Program “Energy System Design”.
% \begin{itemize}
%     \item Wrap Up:
%     \begin{itemize}
%         \item Enhance probabilistic forecasting by incorporating statistical knowledge into the deep learning process. That is applicable to time series forecasting tasks that have a calendar driven periodicity
%         \item Clearly improves the forecasting results
%     \end{itemize}
%     \item Future Work:
%     \begin{itemize}
%         \item Use copula to avoid the assumption that trend, colorful noise and variance are independent.
%         \item Apply that approach on multivariate time series forecasting.
%     \end{itemize}
% \end{itemize}

\appendix


%% \bibliographystyle{elsarticle-num} 

\bibliographystyle{elsarticle-num} 
\bibliography{probpnn}

%\begin{thebibliography}{00}


%\bibitem{}

%\end{thebibliography}
\end{document}